\documentclass[10pt,twocolumn,letterpaper]{article}

\usepackage{iccv}              

\definecolor{iccvblue}{rgb}{0.21,0.49,0.74}

\def\paperID{xxx} 
\def\confName{ICCV}
\def\confYear{2025}

\usepackage{graphicx}         
\usepackage{float}            
\usepackage{caption}          
\usepackage{adjustbox}        
\usepackage{booktabs}         
\usepackage{multirow,array}   

\usepackage{algorithm}
\usepackage{algpseudocode}

\usepackage{nicefrac}         
\usepackage{microtype}        
\usepackage{lipsum}           
\usepackage{verbatim}         
\usepackage{wrapfig}          

\usepackage{hyperref}

\algnewcommand{\Inputs}[1]{%
  \State \textbf{Inputs:}
  \Statex \hspace*{\algorithmicindent}\parbox[t]{.8\linewidth}{\raggedright #1}
}
\algnewcommand{\Outputs}[1]{%
  \State \textbf{Outputs:}
  \Statex \hspace*{\algorithmicindent}\parbox[t]{.8\linewidth}{\raggedright #1}
}
\definecolor{mygreen}{HTML}{5EBE8F}
\definecolor{mean_green}{HTML}{2CA02C}
\definecolor{median_orange}{HTML}{FF973B}

\title{Refining Skewed Perceptions in Vision-Language Contrastive Models through Visual Representations}

\author{Haocheng Dai\\
University of Utah\\
{\tt\small haocheng.dai@utah.edu}
\and
Sarang Joshi\\
University of Utah\\
{\tt\small sarang.joshi@utah.edu}
}

\begin{document}
\maketitle
\begin{abstract}
Large vision-language contrastive models (VLCMs), such as CLIP, have become foundational, demonstrating remarkable success across a variety of downstream tasks. Despite their advantages, these models, akin to other foundational systems, inherit biases from the disproportionate distribution of real-world data, leading to misconceptions about the actual environment. Prevalent datasets like ImageNet are often riddled with non-causal, spurious correlations that can diminish VLCM performance in scenarios where these contextual elements are absent. This study presents an investigation into how a simple linear probe can effectively distill task-specific core features from CLIP's embedding for downstream applications. Our analysis reveals that the CLIP text representations are often tainted by spurious correlations, inherited in the biased pre-training dataset. Empirical evidence suggests that relying on visual representations from CLIP, as opposed to text embedding, is more effective to refine the skewed perceptions in VLCMs, emphasizing the superior utility of visual representations in overcoming embedded biases. Our code can be found \href{https://github.com/aarentai/Refining-Skewed-Perception}{here}.
\end{abstract}
\vspace{-20pt}

\section{Introduction}
\label{sec:intro}
Vision-language contrastive models (VLCMs), a class of multimodal artificial intelligence systems, seamlessly bridge the gap between visual perception and natural language understanding, providing users with a more intuitive way to leverage artificial intelligence for solving daily problems. The synergy between visual and linguistic data has significant implications for various applications, including image generation, image captioning, cross-modal retrieval, and visual question answering.

VLCMs like Contrastive Language-Image Pre-training (CLIP)~\citep{radford2021learning} have set new benchmarks across various tasks by contrastively matching semantically closest image and text pairs. However, due to the disproportionate distribution embedded in real-world datasets like ImageNet~\citep{deng2009imagenet} or LAION~\citep{schuhmann2021laion}, pre-trained VLCMs inherently acquire biases from these large-scale datasets. This phenomenon, known as spurious correlation, refers to patterns that correlate the target class with non-causal contextual elements. For instance, a vision model may classify cows correctly but fail when cows appear outside the typical grassland background, revealing grass as a shortcut predictor for cow~\citep{beery2018recognition}. Similarly, BERT's~\citep{devlin2018bert} peak performance on the argument reasoning comprehension task is largely due to exploiting spurious statistical cues in the dataset, like the negation word ``not''~\citep{niven2019probing}.

In this work, we investigate the spurious correlations embedded in foundational VLCMs like CLIP and aim to answer the following questions: 1) Does CLIP rely on non-causal ``background'' features in its decision-making process? If so, how? 2) Is a linear probe sufficient to distill task-specific core features from CLIP's image embeddings? 3) Can language prompts help us to remove the spurious features for specific tasks? 4) Besides language, can images help to refine the skewed perception in CLIP visual representations for more reliable downstream tasks?

To answer these questions, we conduct various experiments. First, we assess CLIP's zero-shot learning performance on the widely used Waterbirds dataset~\citep{sagawa2019distributionally} before and after removing the ``background'' context. Next, we explore the expressiveness limits of CLIP embeddings by performing various classification tasks on the CelebA~\citep{liu2018large} dataset using only linear probing to see if the embeddings capture nuanced features. To examine the practicality of zero-shot classification in the presence of spurious correlations, we evaluate the degree of contamination in CLIP's text representations due to biased pre-training data through extensive statistical analysis. 
Lastly, we show CLIP's visual representation's ability to distill core features using the experimental \texttt{VisualDistiller} framework.

In summary, in this work, we provide a deeper understanding of the strengths and limitations of CLIP’s visual and textual representations rather than purely benchmarking spurious correlation mitigation methods. Our key findings include:
\begin{itemize}
    \item[--] We show that VLCMs like CLIP rely on non-causal spurious features for decision-making, yet linear probing is sufficient to extract key features for various downstream tasks.
    \item[--] We find that CLIP's text embeddings are contaminated by diverse elements, making text embeddings impractical for debiasing the model.
    \item[--] We demonstrate that using visual embeddings from CLIP to distill visual representations is highly effective. The debiased features achieve excellent performance in group accuracy comparable to state-of-the-art supervised methods~\cite{kirichenko2022last}. This knowledge can inform the development of fairer and more robust machine learning models, such as unbiased image retrieval algorithms and image generation leveraging CLIP.
\end{itemize}

\section{Related Work}
\paragraph{Mitigating Spurious Correlations in Uni-modality Models.} Deep learning frameworks frequently exhibit uneven performance across various groups due to spurious correlations, resulting in notably lower test accuracy for minority groups compared to majority groups. This issue contrasts with the training phase, where both groups generally achieve more balanced training accuracy~\citep{sagawa2019distributionally,geirhos2020shortcut}. \citep{geirhos2020shortcut,shah2020pitfalls,hermann2020shapes} highlight that neural networks are prone to a simplicity bias, often emphasizing trivial spurious features while neglecting the essential core features.

To address these challenges, substantial research has been dedicated to enhancing robustness against spurious correlations. When group labels are available, strategies such as class balancing~\citep{he2009learning,cui2019class}, importance weighting~\citep{shimodaira2000improving,byrd2019effect}, robust optimization~\citep{sagawa2019distributionally,kirichenko2022last,izmailov2022feature}, and contrastive learning~\citep{taghanaki2021robust} have been developed to ensure balanced training across different group sizes. For example, deep feature reweighting (DFR)~\citep{kirichenko2022last} solves this by retraining the last linear layer of the model (e.g. ResNet) by up-weighting the underrepresented samples in the loss function, ensuring better handling of minority groups. In scenarios where group labels are unavailable, a common approach involves initially training an auxiliary model using empirical risk minimization (ERM). The predictions from this model are then used to infer group information, which in turn guides the training of a more robust second model. This robust model is typically trained using techniques such as sample balancing~\citep{liu2021just,nam2020learning}, or contrastive learning~\citep{zhang2022correct,zhang2022contrastive,yang2023mitigating} with the inferred group labels.

\paragraph{Enhancing Group Robustness in VLCMs.} VLCMs have gained increased popularity for their ability to perceive the world through multiple modalities. Previous research has sought to enhance the robustness of vision classifiers by incorporating language features, using techniques such as attention maps~\citep{petryk2022guiding} and modifications to feature attributes~\citep{zhang2023diagnosing}. Significant advancements~\cite{yang2023mitigating, zhang2022contrastive} have been made in developing pre-trained multimodal models resistant to spurious correlations. For instance, \citep{zhang2022contrastive} proposes a novel contrastive adapter that, when combined with transfer learning, improves group robustness. However, this method does not always lead to better results, especially for specific downstream applications. Conversely, \citep{yang2023mitigating} pioneers a fine-tuning strategy specifically designed to address spurious correlations with group labels in pre-trained multimodal models. \citep{chuang2023debiasing} addresses VLCMs' bias in zero-shot classification by projecting out biased directions in the text embeddings. \citep{yuksekgonul2022and} proposes composition-aware hard negatives during training, which improves the model's ability to understand attributes, relations, and order significantly, leading to better performance on tasks that require compositional understanding. Unlike these approaches, our objective is to investigate the inherently skewed perception embedded in all text embeddings (including target class text and spurious attribute text) and explore the possibilities of using visual representations instead to distill the task-specific core features from VLCMs like CLIP for downstream tasks, without the need for group annotations. 

\section{Preliminaries}
\label{sec:pre}
\paragraph{Notations.} In this study, we explore the spurious features inherent in the CLIP visual representations and assess their impact on classification performance. For a given classification task, we have $N$ samples $\{(\mathbf{x}_i, \mathbf{y}_i, a_i, g_i)\}^N_{i=1}$, where $\mathbf{x}_i\in \mathcal{X}$ represents the input features, $\mathbf{y}_i\in \mathcal{Y}$ as the class labels, $a_i\in\mathcal{A}$ as the spurious attributes, and $g_i\in\mathcal{G}=\mathcal{Y}\times\mathcal{A}$ as the group labels. We examine scenarios of distribution shifts occurring between samples across different groups but from the same class. In the Waterbirds dataset~\citep{sagawa2019distributionally}, we define $\mathcal{Y}=$\{landbird, waterbird\}, $\mathcal{A}=$\{land background, water background\}, and $\mathcal{G}=$\{landbird on land ($\mathcal{G}_0$), landbird on water ($\mathcal{G}_1$), waterbird on land ($\mathcal{G}_2$), waterbird on water ($\mathcal{G}_3$)\}. Notably, $\mathcal{G}_1$ and $\mathcal{G}_2$ are the minority groups (fewer training samples), whereas $\mathcal{G}_0$ and $\mathcal{G}_3$ are the majority groups. In the default CelebA dataset~\citep{liu2018large}, the categories are $\mathcal{Y}=$\{non-blond hair, blond hair\}, $\mathcal{A}=$\{female, male\}, and $\mathcal{G}=$\{non-blond hair female ($\mathcal{G}_0$), blond hair female ($\mathcal{G}_1$), non-blond hair male ($\mathcal{G}_2$), blond hair male ($\mathcal{G}_3$)\}. With regard to CelebA, $\mathcal{G}_0, \mathcal{G}_1, \mathcal{G}_2$ are the majority groups, with $\mathcal{G}_3$ being the minority group.
\vspace{-10pt}

\paragraph{Objective.} The training process involves samples $(\mathbf{x}_i, \mathbf{y}_i, a_i, g_i)$ drawn from an unknown joint distribution $P$. We denote $P_g$ as the distribution conditioned on group $g$ for any $g \in \mathcal{G}$. The goal of ERM is to minimize the average classification error using a classifier $f_{\theta}:\mathcal{X}\rightarrow \mathcal{Y}$, described mathematically as:
\begin{equation}
    \mathcal{L}_{\operatorname{avg}}(f_{\theta})=E_{(\mathbf{x},\mathbf{y},a,g)\sim P} [l(f_{\theta}(\mathbf{x}), \mathbf{y})],
\end{equation}
where $l$ is the loss function. To achieve robustness across groups, one aims to minimize the worst-group error:
\begin{equation}
    \mathcal{L}_{\operatorname{wg}}(f_{\theta})=\max_{g\in \mathcal{G}}E_{(\mathbf{x},\mathbf{y},a,g)\sim P_g}[l(f_{\theta}(\mathbf{x}), \mathbf{y})].
\end{equation}

\section{A Linear Probe Can Achieve Optimal Task Performance} 
In modern machine learning system design, the goal is often to enhance foundational models with specialized modules for specific downstream tasks. For classification tasks in particular, the ideal is to utilize the same representations derived from a pre-trained model across various classification challenges. Given the nature of spurious correlations --- where a feature deemed spurious for one task may be essential for another --- we expect the VLCMs to capture a broad spectrum of nuanced visual information and remove the spurious feature by specialized modules. This section delves into the presence of spurious correlations within VLCMs and explores whether a simple linear probe can deliver optimal performance.

\subsection{Unraveling Spurious Correlations in VLCMs}

\begin{figure}
    \centering
    \includegraphics[height=3.5cm, keepaspectratio]{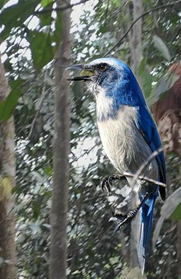}
    \includegraphics[height=3.5cm, keepaspectratio]{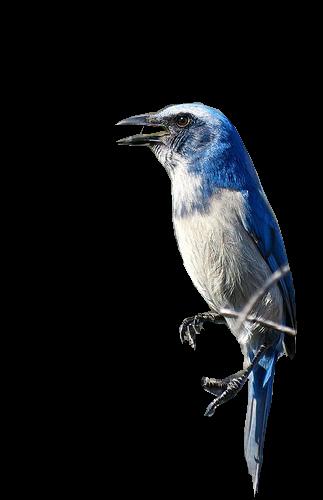}
    \caption{Waterbird samples with background (left) and without background (right). The background is erased via masks available in the dataset.}
    \label{fig:example_fg_only}
    \vspace{-10pt}
\end{figure}

We first examine the presence of spurious correlations in CLIP by evaluating zero-shot binary classification performance on the Waterbirds dataset. Using CLIP’s text and image encoders, we compute the cosine similarity between image features and textual labels (``a photo of a landbird'' and ``a photo of a waterbird''). To assess the impact of spurious correlations, we compare classification performance on the original dataset (with natural backgrounds) and a modified version where backgrounds are removed using segmentation masks (Figure~\ref{fig:example_fg_only}).

\begin{figure*}[t]
    \centering
    \includegraphics[height=4.1cm, keepaspectratio]{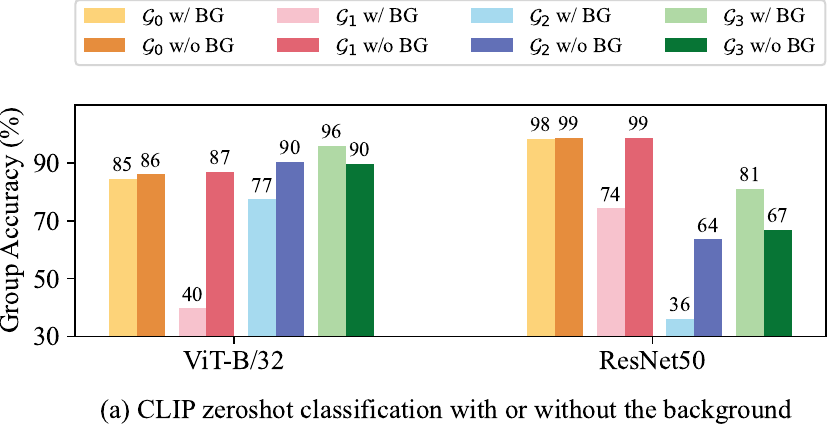}
    \includegraphics[height=4.1cm, keepaspectratio]{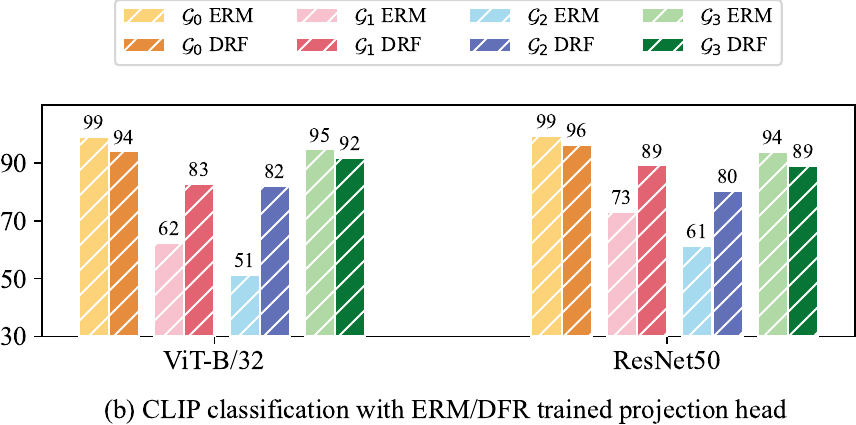}
    \caption{(a) Group accuracy change of CLIP zero-shot classification before and after removing the background on the Waterbirds dataset; (b) Group accuracy change of CLIP classification with ERM and DFR-trained projection head. $\mathcal{G}_0$: landbird on land, $\mathcal{G}_1$: landbird on water, $\mathcal{G}_2$: waterbird on land, $\mathcal{G}_3$: waterbird on water. Notably, $\mathcal{G}_1$ and $\mathcal{G}_2$ are the minority groups (fewer training samples), whereas $\mathcal{G}_0$ and $\mathcal{G}_3$ are the majority groups. BG: background. ``a photo of a landbird/waterbird'' is used as the zero-shot classification prompts. }
    \label{fig:group_accuracy_change_wb}
    \vspace{-10pt}
\end{figure*}


As shown in Figure~\ref{fig:group_accuracy_change_wb}(a), accuracy disparities emerge between majority and minority groups when using the original dataset. Removing backgrounds improves the classification accuracy of minority groups ($\mathcal{G}_1$, $\mathcal{G}_2$), suggesting that background cues previously misled the model. Meanwhile, accuracy drops for the majority group $\mathcal{G}_3$, indicating a stronger reliance on background features for waterbird-on-water samples, aligning with prior work findings \cite{yang2023mitigating,wang2025sober}. These results reinforce that CLIP representations encode spurious features, which can hinder classification performance. This raises a key question: can CLIP-based models achieve robust classification despite these spurious correlations?

\subsection{Assessing the Expressiveness of CLIP's Visual Representations under Linear Probing}
In order to see the upper limit of CLIP visual representation with linear transformation, we applied DFR~\citep{kirichenko2022last}, a state-of-the-art method that retrains the last linear layer of the image classifiers (e.g. ResNet) by up-weighting the underrepresented samples in the loss function, to Waterbirds dataset. Following DFR's implementation, a linear layer is attached to the CLIP image encoder to facilitate binary classification, with updates restricted solely to the weights of the linear layer. As a supervised method (knowing the group label of each sample), DFR usually signifies the peak performance that a linear layer can attain and reduces the impact of spurious attributes without altering the primary network. High accuracy in these groups suggests that the standard CLIP image encoder successfully captures essential task-related features. Figure~\ref{fig:group_accuracy_change_wb}(b) demonstrates the peak classification performance across Waterbirds groups with a CLIP image encoder and a linear projection head. Under DFR, the worst-group accuracy (WGA) improves to 82\% (ViT-B/32) and 80\% (ResNet), exceeding the performance of CLIP model fine-tuned for mitigating spurious correlation~\cite{yang2023mitigating}. This highlights that CLIP representations effectively encode the essential information needed for the original classification task.

\begin{figure*}[t]
    \centering
    \includegraphics[width=0.9\textwidth]{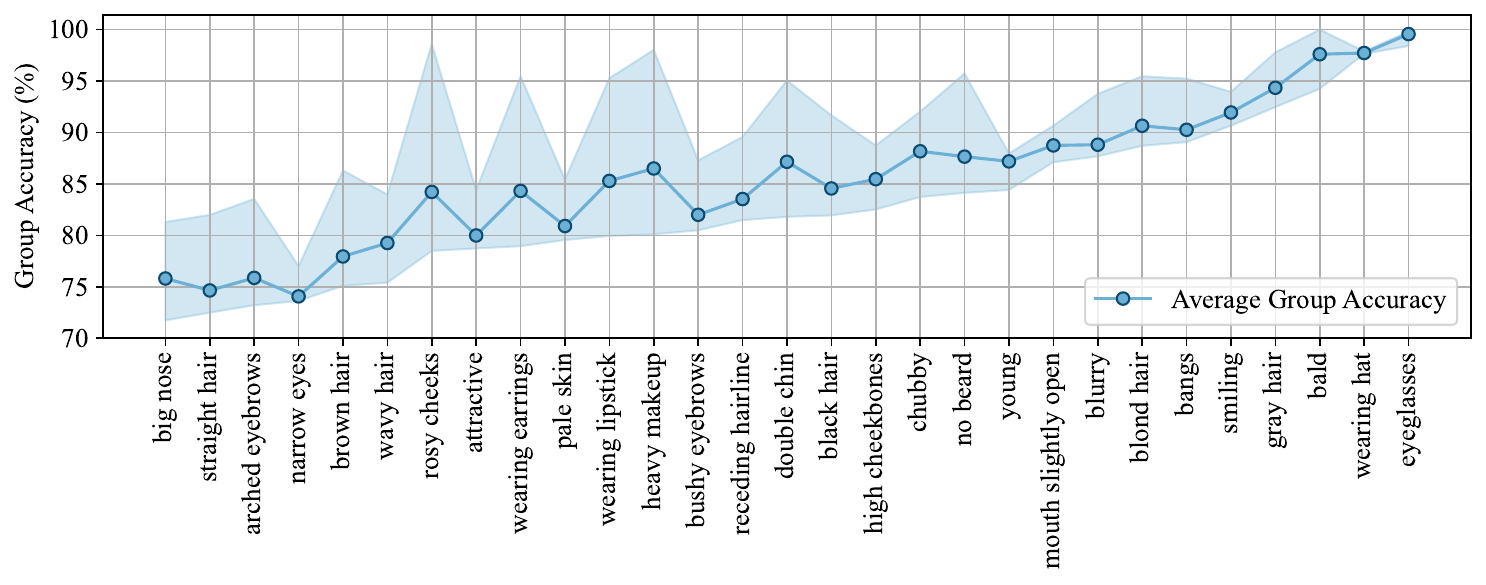}
    \caption{Group accuracy on classifying different CelebA attributes spuriously correlated with gender via training linear probe attached to CLIP image encoder (ViT-L/14). The upper and lower bounds of the shading area stand for best and worst group accuracy. The average group accuracy is a weighted average determined by the size of each group.}
    \label{fig:celeba-other-attr}
    \vspace{-10pt}
\end{figure*}

To further validate the richness of CLIP's visual representation, in Figure~\ref{fig:celeba-other-attr}, we applied DFR to 29 attribute classification challenges on the CelebA dataset, where each attribute demonstrated a gender-biased distribution. The size of each group w.r.t. each attribute can be found in supplementary material. Likewise, they also involve four groups based on gender and attribute presence: female without \texttt{[attribute]} ($\mathcal{G}_0$), male without \texttt{[attribute]} ($\mathcal{G}_1$), female with \texttt{[attribute]} ($\mathcal{G}_2$), and male with \texttt{[attribute]} ($\mathcal{G}_3$). 

Figure~\ref{fig:celeba-other-attr} showcases the outcomes for various binary attribute classifications with a CLIP ViT-L/14 image encoder and a linear projection head, sorted by ascending average group accuracy (a weighted average determined by the size of each group). The spectrum of attributes ranged from subtle features like straight hair and narrow eyes to more overt characteristics such as eyeglasses, baldness, and hats. Notably, the DFR approach strategy enabled a majority of the attributes --- over 25 out of 29 --- to achieve more than 75\% WGA, with more than 23 attribute classification tasks surpassing 80\% average group accuracy, and even outperforming calibrated fine-tuned CLIP~\cite{you2024calibrating}. The five attributes with the highest accuracies exceeded 95\% in both the worst and average group accuracy measures. These results underscore how fine-grained the CLIP’s visual representations are, capturing a comprehensive spectrum of visual information, including subtle features that are typically challenging for human perception. \textbf{Finding 1.} Visual representations learned by CLIP are adept at extracting nuanced features within images for various tasks by linear transformation.

\begin{figure}
    \centering
    \includegraphics[width=\linewidth]{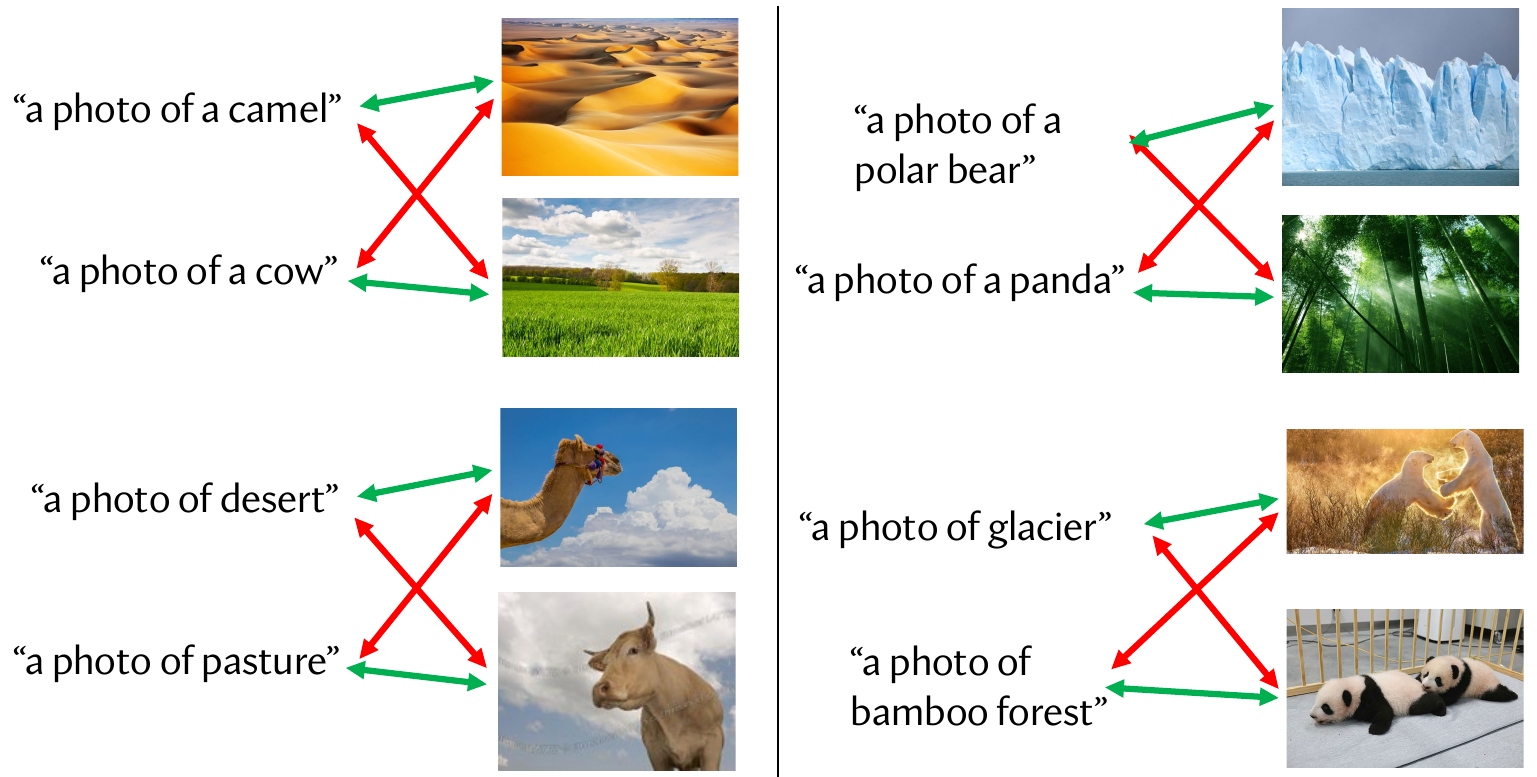}
    \caption{An illustration of text-image pair similarity comparison. We make sure that all the images used in the calculation are absent of the object mentioned in the text prompt in the calculation. The \textcolor{red}{red} double-headed arrow indicates a longer distance (lower similarity) measured by cosine similarity between embeddings.}
    \label{fig:cosine_similarity_illustration}
    \vspace{-20pt}
\end{figure}

\begin{figure*}[t]
    \centering
    \includegraphics[height=3.8cm, keepaspectratio]{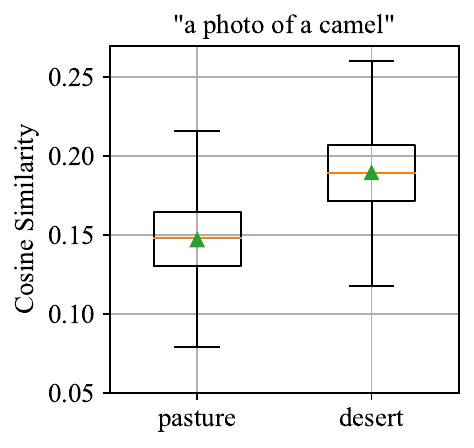}
    \includegraphics[height=3.8cm, keepaspectratio]{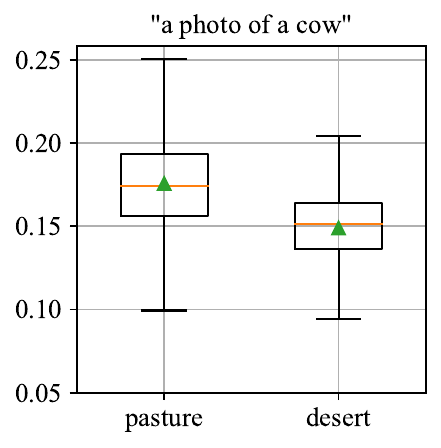}
    \includegraphics[height=3.8cm, keepaspectratio]{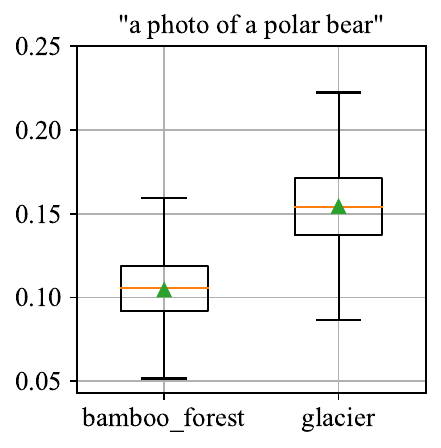}
    \includegraphics[height=3.8cm, keepaspectratio]{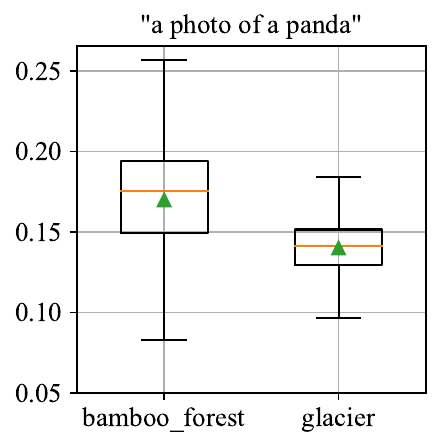}
    \includegraphics[height=3.8cm, keepaspectratio]{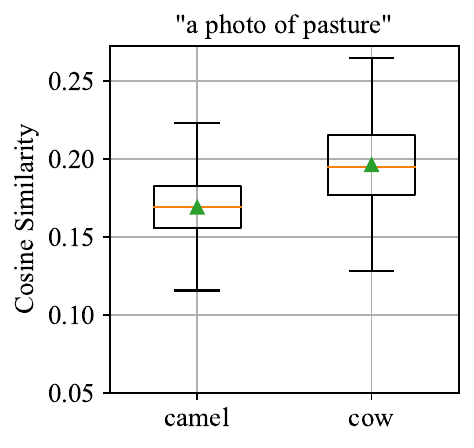}
    \includegraphics[height=3.8cm, keepaspectratio]{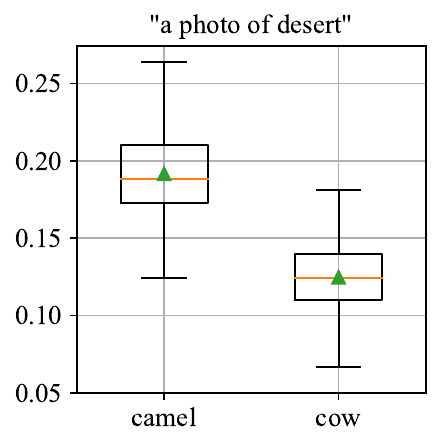}
    \includegraphics[height=3.8cm, keepaspectratio]{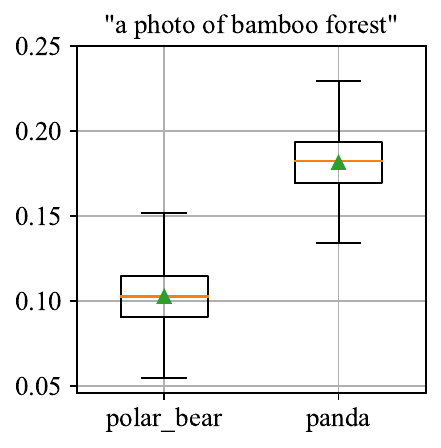}
    \includegraphics[height=3.8cm, keepaspectratio]{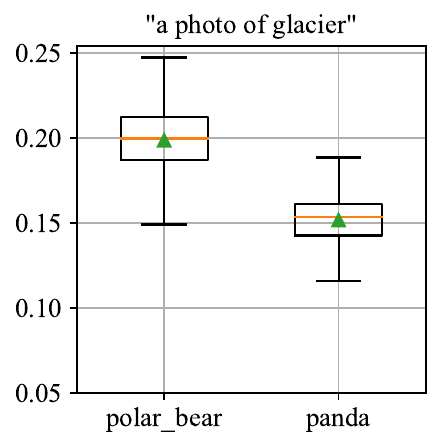}
    \caption{Cosine similarities between spuriously correlated text prompt (boxplot title) and images (x-axis labels) pair. \textcolor{mean_green}{Mean} and \textcolor{median_orange}{median} values of the cosine similarities are denoted by \textcolor{mean_green}{green triangle} and \textcolor{median_orange}{orange line}. For each category of images (camel/cow/bear/panda/pasture/desert/forest/glacier), 800 to 1,000 images are collected for evaluation. For a more intuitive illustration, see Figure~\ref{fig:cosine_similarity_illustration}.}
    \label{fig:cosine-similarity}
    \vspace{-10pt}
\end{figure*}

\section{Language Representations are More Noisy than Expected}
VLCMs like CLIP prevail partly because of their capability to perform zero-shot inferences guided by intuitive language cues. As demonstrated in the previous section, techniques such as DFR guide us toward identifying an optimal linear probe that can effectively discern variations in core features. Similarly, in zero-shot learning, the linear probe is formed by concatenating text representations. This section explores the biases inherent in zero-shot classification prompts.

VLCMs are trained to align the representations of images with their corresponding captions via cosine similarities. Ideally, one might expect the representation of ``a photo of a dog'' to solely encapsulate the dog's key features without incorporating ambient elements like lawns. However, the examination of real-world data reveals a spurious correlation where dog images are typically associated with outdoor environments, and cat images are often taken indoors. We hypothesize that these contextual features are inevitably embedded in the CLIP text representations.

To test this hypothesis, we examined various prompts and corresponding image pairs, calculating the cosine similarity between them. In Figure~\ref{fig:cosine_similarity_illustration}, we evaluated pairs like (``a photo of a camel''/``a photo of a cow'', desert/pasture images), (``a photo of a polar bear''/``a photo of a panda'', glacier/bamboo forest images), and conversely (``a photo of pasture''/``a photo of desert'', camel/cow images), (``a photo of bamboo forest''/``a photo of glacier'', polar bear/panda images), with ensuring the images tested here did not contain the objects mentioned in the prompts. This methodology helps quantify the extent of spurious features embedded in text representations.

Figure~\ref{fig:cosine-similarity} shows the result. Notably, the cosine similarity distributions, indicated by the \textcolor{mean_green}{mean (green triangle)} and \textcolor{median_orange}{median (orange line)}, reveal strong correlations—for example, the prompt ``a photo of a camel'' with camel-free desert images (top left in Figure~\ref{fig:cosine_similarity_illustration}) and the prompt ``a photo of a cow'' with cow-free pasture images (top left in Figure~\ref{fig:cosine_similarity_illustration}). This pattern is consistent across various tested pairs, underscoring the substantial presence of context-related features in CLIP text representations that are not explicitly present in the prompts. For the prompt and image pair with less pronounced correlations, like ``a photo of a dog'' to forest and desert, we did not observe the same level of disparity in mean and median values. For disparity on additional text and image pairs, see supplementary materials.

\textbf{Finding 2.} Using text representations for zero-shot classification or debiasing with the representations from spurious attribute prompts~\citep{chuang2023debiasing} could lead to unexpected outcomes, due to the embedded non-target features.

\section{Visual Representations Can Refine the Skewed Perception in VLCMs}
\begin{figure*}[h]
    \centering
    \includegraphics[width=0.75\textwidth]{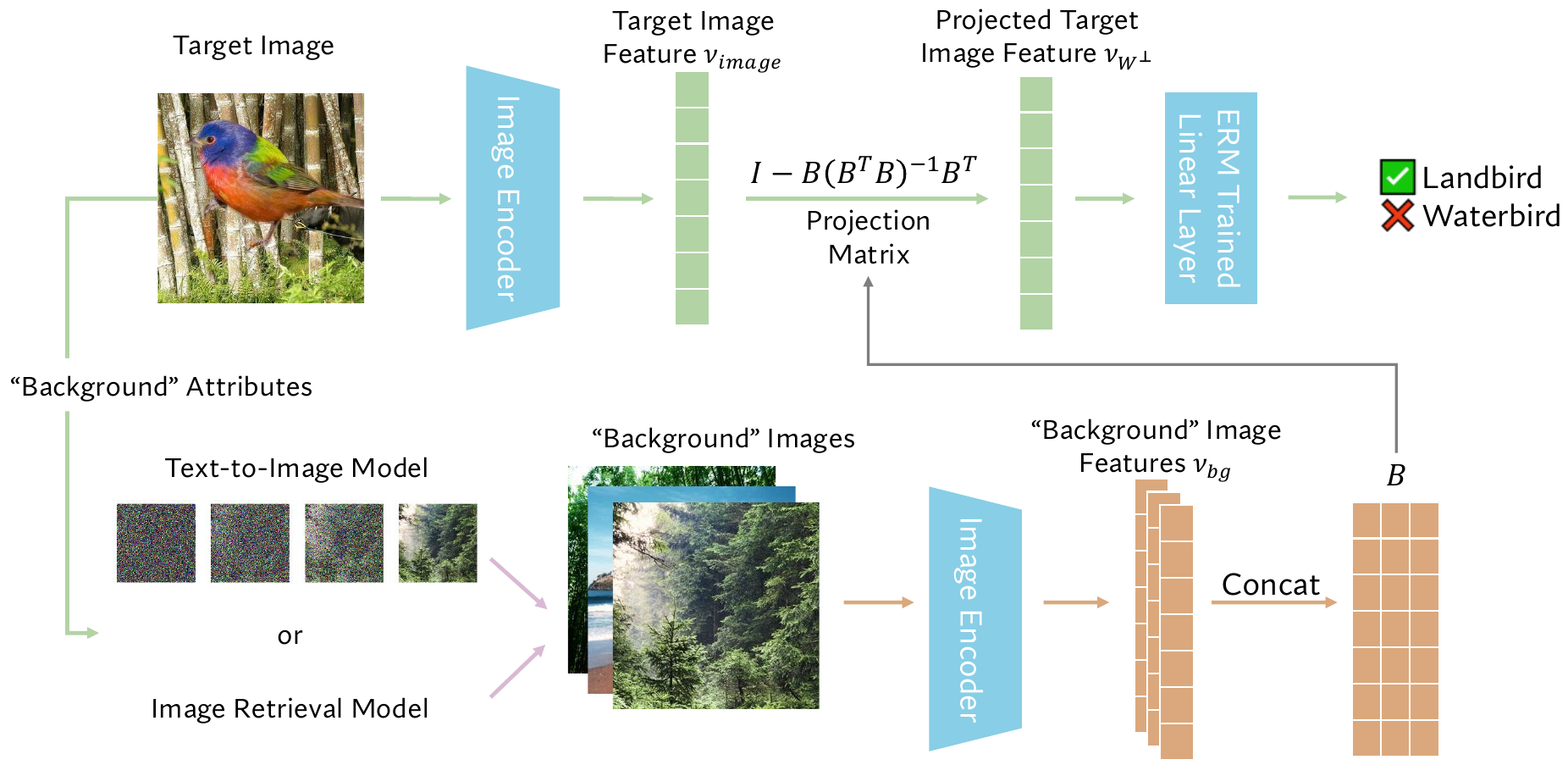}
    \caption{The experimental \texttt{VisualDistiller} framework. }
    \label{fig:visual_distiller}
    \vspace{-10pt}
\end{figure*}

Due to the noisy nature of the text embeddings from CLIP, we wonder if we can use images to construct an optimal linear probe. The availability of background images in the Waterbirds dataset inspires us to explore whether we can use the background images to remove non-core features from the representations of the original Waterbirds images. In Figure~\ref{fig:visual_distiller}, we demonstrate an experimental framework \texttt{VisualDistiller}, using only ERM along with a few inference and projection steps ---making it significantly more lightweight than retraining the entire CLIP framework. We opted for an ERM linear probe (trained by only one epoch, without knowing the group labels of training samples) over zero-shot text classification due to the existence of context-related features in text representations, which compromise classification reliability. 

The process is as follows: After encoding the target image via CLIP, we obtained the target image feature $\mathbf{v}_{\operatorname{image}}\in\mathbb{R}^n$. Prior to projection, we aim to isolate the ``background'' component from $\mathbf{v}_{\operatorname{image}}$. We model this as a linear problem by constructing a subspace $W$ in $\mathbb{R}^n$, spanned by $m$ ``background'' vectors $\mathbf{v}_{\operatorname{bg}}\in\mathbb{R}^n$. Assuming $\mathbf{v}_{\operatorname{image}} = \mathbf{v}_{W} + \mathbf{v}_{W^\perp}$, where $\mathbf{v}_{W}$ is closest vector to $\mathbf{v}_{\operatorname{image}}$ and $\mathbf{v}_{W^\perp}$ lies in the orthogonal complement $W^\perp$, we define $B$ as an $n \times m$ matrix of linearly independent columns ($\mathbf{v}_{\operatorname{bg}}$) and $W = \operatorname{Col}(B)$. The orthogonal component $\mathbf{v}_{W^\perp}$ is calculated as:
\begin{equation}
    \mathbf{v}_{W^\perp} = (I - B(B^T B)^{-1} B^T) \mathbf{v}_{\operatorname{image}},\label{eq:proj}
\end{equation}
(see proof in supplementary materials). $\mathbf{v}_{W^\perp}$ is then processed through the ERM-trained linear probe to produce the final classification result. For simplicity and ease of implementation, we assume Euclidean orthogonality, proven effective in~\cite{krishna2023imaginator,materzynska2022disentangling}. Using non-Euclidean orthogonality --- specifically, employing the inverse of the covariance matrix in the neighborhood space as the metric for inner product calculations --- did not yield substantial improvement.

\begin{figure*}[h]
    \centering
    \includegraphics[width=1\textwidth]{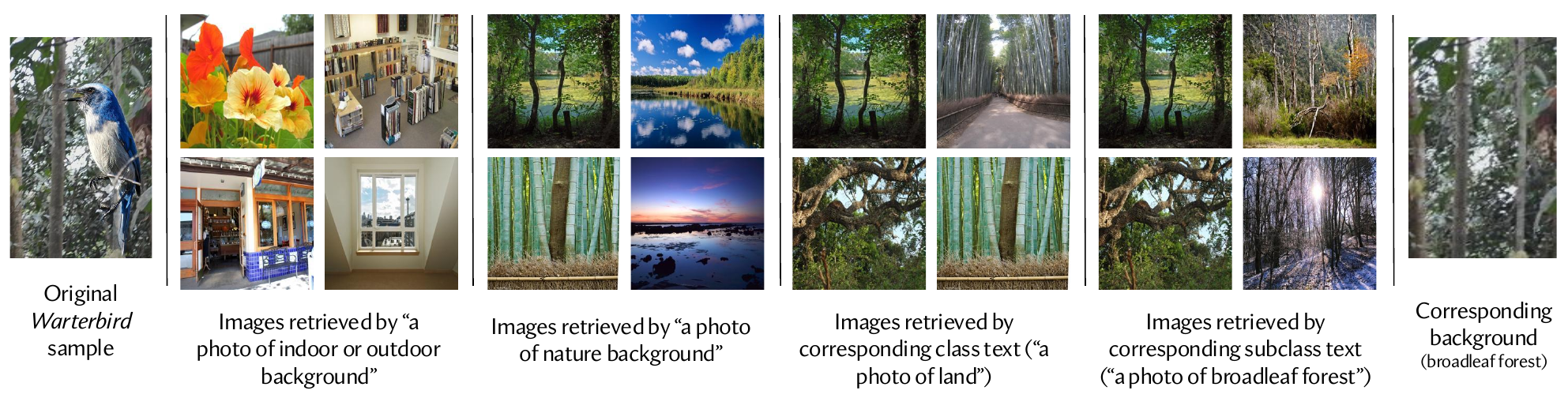}
    \caption{Examples of background images with varying levels of semantic similarity to the original Waterbird sample's background.}
    \label{fig:backgound_image_example}
\end{figure*}

The definition of a ``background'' image varies with the dataset. For the Waterbirds dataset, artificially created with images from Places~\citep{zhou2016places} and CUB~\citep{WahCUB_200_2011}, we defined a range of ``background'' conditions from loosely related (retrieved\footnote{Here, ``retrieved'' refers to the process of encoding all images in the Places dataset using the CLIP ViT-L/14 image encoder and then identifying the top 50 images whose embeddings are most similar to the text embedding using K-nearest neighbors.} images from Places using ``a photo of indoor or outdoor background'') to more strictly related (retrieved images from Places using ``a photo of nature background'') to corresponding backgrounds used in Waterbirds. Figure~\ref{fig:backgound_image_example} shows examples of background images with varying levels of semantic similarity, which is used in Table~\ref{tab:waterbirds}. In Table~\ref{tab:waterbirds}, we demonstrate that the experiments using semantically more similar ``background'' images yield higher WGA in Waterbirds. Besides, transitioning to an ERM-trained linear probe enhances WGA further by focusing more sharply on core features, unlike the ``contaminated'' text representations from CLIP. Both supervised (knowing the corresponding ``background'' category, denoted by $\mathparagraph$ in Table~\ref{tab:waterbirds}) and unsupervised (without knowing the corresponding ``background'' category, denoted by $\dagger$) projections were explored, with the supervised setup serving to illustrate the upper limit of this approach, rather than its practical applicability. Increasing the number of ``background'' vectors generally improves WGA, but with diminishing returns. The \texttt{VisualDistiller} can achieve the WGA of 82.40\% without knowing the background image category (20 random images from nature, ViT), only a few points from supervised DFR's 85.67\% performance. We also observe a minor decrease in the average accuracy of our proposed experiments compared to ERM without projection on the original Waterbirds dataset. This is expected, and widely exists in similar framework~\cite{yang2023mitigating,you2024calibrating}, as ERM prioritizes overall accuracy at the expense of minority group performance --- the main optimization subject of the work.

\begin{table*}[h]
\caption{\textbf{Group accuracy by zero-shot/ERM/DFR classification on Waterbirds} dataset across different CLIP backbones and projection operations. Corresponding class (subclass) text refers to ``a photo of land/waterbody'' (``a photo of ocean/lake/broadleaf forest/bamboo forest''). The corresponding background is retrieved from the Waterbirds metadata file. Figure~\ref{fig:backgound_image_example} explains the types in column \textit{``Background'' Vector Source} column intuitively. WGA: worst group accuracy; Avg: average group accuracy. $\dagger$: unsupervised method; $\mathparagraph$: supervised method.}
\centering
\label{tab:waterbirds}
\scalebox{0.83}{
\begin{tabular}{clccccc}
\toprule
\multicolumn{1}{l}{}                                                       & \textbf{}                                                                                   & \multicolumn{1}{l}{\textbf{}}                                        & \multicolumn{2}{c}{\textbf{CLIP ViT}}                   & \multicolumn{2}{c}{\textbf{CLIP ResNet}}                \\\cmidrule(lr){1-3}\cmidrule(lr){4-5}\cmidrule(lr){6-7}
\textbf{\begin{tabular}[c]{@{}c@{}}Projection \\ Head Source\end{tabular}} & \textbf{\begin{tabular}[c]{@{}l@{}}``Background''\\ Vector Source\end{tabular}} & \textbf{\begin{tabular}[c]{@{}c@{}}``Background'' \\ Vector \#\end{tabular}} & \multicolumn{1}{c}{\textbf{WGA}$\uparrow$}                              & \multicolumn{1}{c}{\textbf{Avg}$\uparrow$} & \multicolumn{1}{c}{\textbf{WGA}$\uparrow$}                              & \multicolumn{1}{c}{\textbf{Avg}$\uparrow$} \\\midrule
                                                                           & $\dagger$ no projection                                                                               & n/a                                                                  & \cellcolor[HTML]{EEA7A1}35.67\%          & 90.41\%      & \cellcolor[HTML]{EB9790}36.14\%          & 92.89\%      \\\cmidrule(lr){2-7}
                                                                           & $\mathparagraph$ corresponding class text                                                                    & 1                                                                    & \cellcolor[HTML]{E67C73}17.45\%          & 86.31\%      & \cellcolor[HTML]{E67C73}26.64\%          & 92.07\%      \\
                                                                           & $\mathparagraph$ corresponding subclass text                                                                 & 1                                                                    & \cellcolor[HTML]{F3C2BE}46.57\%          & 89.43\%      & \cellcolor[HTML]{EFAFAA}44.24\%          & 93.09\%      \\\cmidrule(lr){2-7}
                                                                           & $\dagger$ an image retrieved by ``a photo of indoor or outdoor background''                                     & 1                                                                    & \cellcolor[HTML]{F1B8B3}42.68\%          & 90.56\%      & \cellcolor[HTML]{F2BBB6}48.29\%          & 90.65\%      \\
                                                                           & $\mathparagraph$ an image retrieved by corresponding class text                                                                 & 1                                                                    & \cellcolor[HTML]{F7D6D3}54.83\%          & 86.95\%      & \cellcolor[HTML]{FCF1F0}66.82\%          & 86.41\%      \\
                                                                           & $\mathparagraph$ an image retrieved by corresponding subclass text                                                              & 1                                                                    & \cellcolor[HTML]{F8DDDB}57.94\%          & 87.51\%      & \cellcolor[HTML]{FFFFFF}71.34\%          & 81.68\%      \\\cmidrule(lr){2-7}
\multirow{-7}{*}{Zero-shot}                                                & $\mathparagraph$ the corresponding background                                                                & 1                                                                    & \cellcolor[HTML]{F7D7D4}55.45\%          & 87.55\%      & \cellcolor[HTML]{B7E2CD}75.23\%          & 87.65\%      \\\midrule
                                                                           & $\dagger$ no projection, original Waterbirds                                                          & n/a                                                                  & \cellcolor[HTML]{FBFEFC}72.27\%          & 97.83\%      & \cellcolor[HTML]{F9E1DF}61.37\%          & 96.62\%      \\\cmidrule(lr){2-7}
                                                                           &                                                                                             & 1                                                                    & \cellcolor[HTML]{FEFAFA}70.09\%          & 96.49\%      & \cellcolor[HTML]{F9E3E1}61.84\%          & 94.92\%      \\
                                                                           &                                                                                             & 3                                                                    & \cellcolor[HTML]{FEFAFA}70.09\%          & 96.31\%      & \cellcolor[HTML]{F9E4E2}62.15\%          & 94.71\%      \\
                                                                           & \multirow{-3}{*}{$\dagger$ images retrieved by ``a photo of indoor or outdoor background''}                    & 10                                                                   & \cellcolor[HTML]{FFFFFF}71.81\%          & 96.06\%      & \cellcolor[HTML]{FAE6E5}63.08\%          & 94.08\%      \\\cmidrule(lr){2-7}
                                                                           &                                                                                 & 1                                                                           & \cellcolor[HTML]{FFFFFF}77.73\%          & 97.33\%      & \cellcolor[HTML]{FCF4F3}62.93\%          & 95.61\%      \\
                                                                           &                                                                                 & 3                                                                           & \cellcolor[HTML]{EBF7F2}78.97\%          & 97.23\%      & \cellcolor[HTML]{FFFFFF}66.20\%          & 94.11\%      \\
                                                                           &                                                                                 & 10                                                                          & \cellcolor[HTML]{C3E7D6}81.46\%          & 96.26\%      & \cellcolor[HTML]{FCEFEE}61.53\%          & 91.17\%      \\
                                                                           & \multirow{-4}{*}{$\dagger$ images retrieved by  ``a photo of nature background''}                                       & 20                                                                          & \cellcolor[HTML]{B4E1CB}82.40\%          & 95.03\%      & \cellcolor[HTML]{FCF3F2}62.77\%          & 90.15\%      \\\cmidrule(lr){2-7}
                                                                           &                                                                                             & 1                                                                    & \cellcolor[HTML]{98D5B7}81.93\%          & 96.20\%      & \cellcolor[HTML]{D7EFE3}73.52\%          & 93.60\%      \\
                                                                           &                                                                                             & 3                                                                    & \cellcolor[HTML]{6BC398}86.29\%          & 95.47\%      & \cellcolor[HTML]{74C79F}78.82\%          & 91.74\%      \\

                                                                           & \multirow{-3}{*}{$\mathparagraph$ images retrieved by corresponding class text}                                                & 10                                                                   & \cellcolor[HTML]{63C092}87.07\%          & 93.45\%      & \cellcolor[HTML]{CEECDD}73.99\%          & 89.86\%      \\\cmidrule(lr){2-7}
                                                                           &                                                                                             & 1                                                                    & \cellcolor[HTML]{7FCCA6}84.27\%          & 95.84\%      & \cellcolor[HTML]{C8E9D9}74.30\%          & 94.09\%      \\
                                                                           &                                                                                             & 3                                                                    & \cellcolor[HTML]{5EBE8F}87.54\%          & 94.15\%      & \cellcolor[HTML]{63C093}79.75\%          & 92.05\%      \\
                                                                           & \multirow{-3}{*}{$\mathparagraph$ images retrieved by corresponding subclass text}                                             & 10                                                                   & \cellcolor[HTML]{5BBD8D}87.85\%          & 93.35\%      & \cellcolor[HTML]{E3F4EB}72.90\%          & 89.82\%      \\\cmidrule(lr){2-7}
                                                                           & $\mathparagraph$ corresponding background                                                                    & n/a                                                                  & \cellcolor[HTML]{57BB8A}88.16\%          & 96.71\%      & \cellcolor[HTML]{6CC499}79.28\%          & 93.83\%      \\\cmidrule(lr){2-7}
\multirow{-16}{*}{ERM}                                                     & $\dagger$ no projection, background removed                                               & n/a                                                                         & \cellcolor[HTML]{57BB8A}91.12\% & 97.69\%      & \cellcolor[HTML]{57BB8A}87.23\% & 96.25\%      \\
\midrule
DFR\cite{kirichenko2022last}                                              & $\mathparagraph$ no projection, original Waterbirds                                                                               & n/a                                                                  & \cellcolor[HTML]{71C69C}85.67\%          & 97.45\%      & \cellcolor[HTML]{57BB8A}80.37\% & 94.19\%     \\\bottomrule
\end{tabular}}
\end{table*}

\begin{figure}[H]
    \centering
    \includegraphics[width=0.85\linewidth]{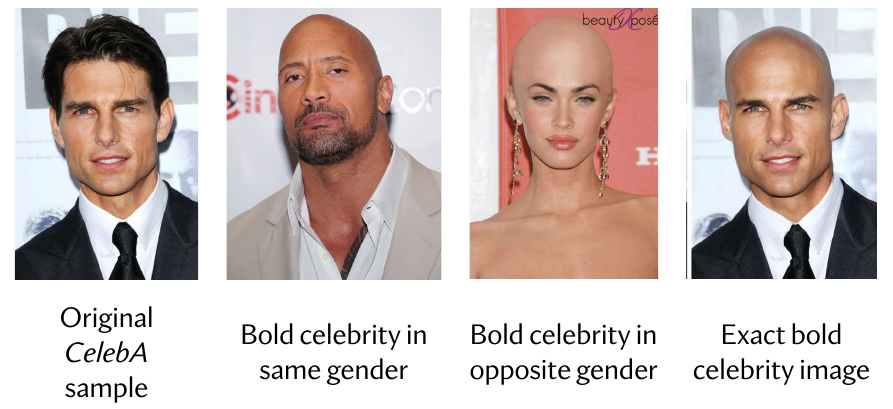}
    \caption{Examples of different ``background'' images in classifying hair color in the CelebA dataset.}
    \label{fig:bold_image_example}
    \vspace{-10pt}
\end{figure}

Additionally, we applied \texttt{VisualDistiller} to the CelebA dataset classifying if the celebrity's hair color is blond. Here, we used retrieved\footnote{Here, ``retrieved'' refers to the process of encoding all images in the CelebA dataset using the CLIP ViT-L/14 image encoder and then identifying the top 50 images whose embeddings are most similar to the query embedding using K-nearest neighbors.} images of celebrities without hair as ``background'' vector source to mitigate non-hair related features. For examples of ``background'' images in this experiment, see Figure~\ref{fig:bold_image_example}. Despite real-world limitations preventing the exact matching of these ``background'' conditions, using a set of bald celebrity images proved effective. In Table~\ref{tab:celeba}, results show significant improvements in WGA with ERM projections, particularly when using gender-matched bald celebrity images. We observed that the WGA on an ERM linear probe escalated from 47.22\%/38.89\% (no projection on ViT/ResNet) to 83.88\%/83.33\% (projecting with a corresponding gender bald image on ViT/ResNet). However, projections using irrelevant or opposite-gender images tended to reduce the WGA gains achieved through gender-matching bald images, highlighting the specificity required for effective ``background'' vector selection. Although using text-based ``background'' vectors assisted in refining the projection, the inherent biases within text representations limited their effectiveness compared to image-based projections. More experiments on other CelebA attributes can be found in Table~\ref{tab:celeba_other_attr}. \textbf{Note that} the proposal of \texttt{VisualDistiller} purely aims to validate the effectiveness of visual representation over the text representation, hence we do not seek to benchmark with other methods like supervised DFR.

\begin{table*}[h]
\caption{\textbf{Group accuracy by ERM/DFR classification (whether the hair color is blond) on CelebA} dataset across different CLIP backbones and projection operations. Corresponding gender (opposite gender/irrelevant) text refers to the prompt ``a photo of a bald male/female celebrity'' (``a photo of a bald female/male celebrity''/``98sa7dyf978yre487fyhs9uihf''). An irrelevant image refers to a Waterbirds photo. Figure~\ref{fig:bold_image_example} explains the types in column \textit{``Background'' Vector Source} column intuitively. WGA: worst group accuracy; Avg: average group accuracy. $\dagger$: unsupervised method; $\mathparagraph$: supervised method.}
\centering
\label{tab:celeba}
\scalebox{0.9}{
\begin{tabular}{clrrrr}
\toprule
\multicolumn{1}{l}{}            & \textbf{}                                                                                                       & \multicolumn{2}{c}{\textbf{CLIP ViT}}                                       & \multicolumn{2}{c}{\textbf{CLIP ResNet}}                                    \\\cmidrule(lr){1-2}\cmidrule(lr){3-4}\cmidrule(lr){5-6}
\textbf{\begin{tabular}[c]{@{}c@{}}Projection \\ Head Source\end{tabular}} & \multicolumn{1}{l}{\textbf{\begin{tabular}[l]{@{}l@{}}``Background''\\ Vector Source\end{tabular}}} & \multicolumn{1}{c}{\textbf{WGA}$\uparrow$}          & \multicolumn{1}{c}{\textbf{Avg}$\uparrow$} & \multicolumn{1}{c}{\textbf{WGA}$\uparrow$}          & \multicolumn{1}{c}{\textbf{Avg}$\uparrow$} \\\midrule
                                & $\dagger$ no projection                                                                                                   & \cellcolor[HTML]{E67C73}47.22\%          & 94.78\%                          & \cellcolor[HTML]{E67C73}38.89\%          & 95.29\%                          \\ \cmidrule(lr){2-6}
                                & $\dagger$ irrelevant text                                                                                                 & \cellcolor[HTML]{F4C9C5}61.67\%          & 93.95\%                          & \cellcolor[HTML]{F2BCB7}50.56\%          & 94.99\%                          \\
                                & $\mathparagraph$ opposite gender text                                                                                            & \cellcolor[HTML]{F4C9C5}61.67\%          & 93.79\%                          & \cellcolor[HTML]{ECA09A}45.56\%          & 94.99\%                          \\
                                & $\mathparagraph$ corresponding gender text                                                                                       & \cellcolor[HTML]{FBECEB}68.33\%          & 93.76\%                          & \cellcolor[HTML]{F3C5C1}52.22\%          & 95.05\%                          \\ \cmidrule(lr){2-6}
                                & $\dagger$ an irrelevant image                                                                                             & \cellcolor[HTML]{F1BAB5}58.89\%          & 93.81\%                          & \cellcolor[HTML]{F7D7D4}55.56\%          & 94.38\%                          \\
                                & $\mathparagraph$ an image retrieved by opposite gender text                                                                                        & \cellcolor[HTML]{F9E3E1}66.67\%          & 85.45\%                          & \cellcolor[HTML]{EBF7F1}66.11\%          & 87.98\%                          \\
                                & $\dagger$ two images retrieved by corresponding and opposite gender text                                                                                         & \cellcolor[HTML]{9ED8BC}79.37\%          & 86.21\%                          & \cellcolor[HTML]{92D3B3}81.11\%          & 87.43\%                          \\
\multirow{-8}{*}{ERM}           & $\mathparagraph$ an image retrieved by corresponding gender text                                                                                    & \cellcolor[HTML]{8CD1AF}83.88\%          & 87.60\%                          & \cellcolor[HTML]{80CCA6}83.33\%          & 87.76\%                          \\\midrule
DFR                             & $\mathparagraph$ no projection                                                                                                   & \cellcolor[HTML]{57BB8A}89.38\% & 90.70\%                          & \cellcolor[HTML]{57BB8A}89.77\% & 91.38\%  \\ \bottomrule                       
\end{tabular}
}
\end{table*}

\begin{table*}[h]
\caption{\textbf{Worst group accuracy by ERM/DFR classification on CelebA} dataset across different projection operations. The corresponding (opposite) gender text refers to the prompt ``a photo of bald male/female celebrity'' (``a photo of bald female/male celebrity''). An irrelevant image refers to a Waterbirds photo. $\dagger$: unsupervised method; $\mathparagraph$: supervised method.}
\centering
\label{tab:celeba_other_attr}
\scalebox{0.9}{
\begin{tabular}{lclc}
\toprule
\textbf{Attributes}          & \textbf{\begin{tabular}[c]{@{}c@{}}Projection \\ Head Source\end{tabular}} & \textbf{\begin{tabular}[c]{@{}l@{}}``Background''\\ Vector Source\end{tabular}} & \textbf{Worst Group Accuracy} \\\midrule
                             &                                                                            & $\dagger$no projection                                                               & \cellcolor[HTML]{DDF2E8}88.24\%  \\
                             &                                                                            & $\dagger$an irrelevant image                                                         & \cellcolor[HTML]{89CFAD}94.61\%  \\
                             &                                                                            & $\mathparagraph$an image retrieved by opposite gender text                                                   & \cellcolor[HTML]{77C8A0}95.97\%  \\
                             &                                                                            & $\dagger$a male and female image                                                     & \cellcolor[HTML]{69C397}96.97\%  \\
                             & \multirow{-5}{*}{ERM}                                                      & $\mathparagraph$an image retrieved by corresponding gender text                                               & \cellcolor[HTML]{85CEAA}94.85\%  \\\cmidrule(lr){2-4}
\multirow{-6}{*}{Black Hair} & DFR                                                                        & $\mathparagraph$no projection                                                               & \cellcolor[HTML]{7DCBA5}95.48\%  \\\midrule
                             &                                                                            & $\dagger$no projection                                                               & \cellcolor[HTML]{F8DDDB}76.84\%  \\
                             &                                                                            & $\dagger$an irrelevant image                                                         & \cellcolor[HTML]{F4C6C2}70.44\%  \\
                             &                                                                            & $\mathparagraph$an image retrieved by opposite gender text                                                   & \cellcolor[HTML]{5CBD8D}97.97\%  \\
                             &                                                                            & $\dagger$a male and female image                                                     & \cellcolor[HTML]{57BB8A}98.29\%  \\
                             & \multirow{-5}{*}{ERM}                                                      & $\mathparagraph$an image retrieved by corresponding gender text                                               & \cellcolor[HTML]{89D0AD}94.56\%  \\\cmidrule(lr){2-4}
\multirow{-6}{*}{Brown Hair} & DFR                                                                        & $\mathparagraph$no projection                                                               & \cellcolor[HTML]{7ECBA5}95.41\%  \\\bottomrule
\end{tabular}}
\end{table*}

\section{Conclusions}
\label{sec:conclusions}
In this study, we explored the capabilities of a CLIP model in manipulating a linear probe from multiple perspectives. We showed that the text prompt representations are often tainted by contextual features embedded within the training data. 
In contrast, visual representations demonstrated greater expressiveness, and targeting specific features within these representations proved highly effective for extracting essential information for downstream tasks. Our straightforward, cost-effective, and potent framework \texttt{VisualDistiller} is intended to generate further insights into the crafting of representations in VLCMs and provide the community with a more comprehensive understanding of the distinct capabilities and limitations of CLIP's visual and textual representations.
    \vspace{-10pt}

\paragraph{Broader Impacts.} Mitigating spurious correlations in machine learning models is essential for building reliable and trustworthy AI. For instance, recognizing that text embeddings contain spurious biases helps us approach their use in image retrieval with caution. In future work, we aim to enhance fairness in image retrieval by leveraging distilled image embeddings as a surrogate for query embeddings, reducing bias in retrieved results. 
\vspace{-10pt}

\newpage
\bibliographystyle{plainnat}
\bibliography{bibliography}

\end{document}